\documentclass[letterpaper, 10 pt, conference]{ieeeconf}  % Comment this line out if you need a4paper

\IEEEoverridecommandlockouts                              % This command is only needed if 
\usepackage{graphics} % for pdf, bitmapped graphics files
\usepackage{epsfig} % for postscript graphics files
\usepackage{times} % assumes new font selection scheme installed
\usepackage{amsmath} % assumes amsmath package installed
\usepackage{amssymb}  % assumes amsmath package installed
\usepackage{caption}
\usepackage{subcaption}
\usepackage{multirow}

\usepackage{booktabs}
\newcommand{\ra}[1]{\renewcommand{\arraystretch}{#1}}

\newcommand{\xd}{\dot{x}}
\newcommand{\xdd}{\ddot{x}}
\usepackage[bookmarks=true]{hyperref}

\usepackage[disable]{todonotes}
\newcommand{\FM}[1]{\todo[inline,color=green!40]{F: #1}}
\newcommand{\GS}[1]{\todo[inline,color=blue!40]{G: #1}}
\newcommand{\AR}[1]{\todo[inline,color=orange!40]{A: #1}}
\title{\LARGE \bf Learning Feedback Terms for Reactive Planning and
  Control} \author{Akshara Rai$^{2,3,*}$, Giovanni Sutanto$^{1,2,*}$,
  Stefan Schaal$^{1,2}$ and Franziska
  Meier$^{1,2}$% <-this % stops a space
  \thanks{$^{*}$both authors contributed equally to this work}%
  \thanks{$^{1}$CLMC lab, University of Southern California, Los
    Angeles, USA.}%
  \thanks{$^{2}$Autonomous Motion Department, MPI-IS, T\"ubingen,
    Germany.}%
  \thanks{$^{3}$Robotics Institute, Carnegie Mellon University,
    Pittsburgh, USA.}%
  \thanks{This research was supported in part by National Science
    Foundation grants IIS-1205249, IIS-1017134, EECS-0926052, the
    Office of Naval Research, the Okawa Foundation, and the
    Max-Planck-Society.}
}

\begin{document}

\maketitle
\thispagestyle{empty}
\pagestyle{empty}

%%%%%%%%%%%%%%%%%%%%%%%%%%%%%%%%%%%%%%%%%%%%%%%%%%%%%%%%%%%%%%%%%%%%%%%%%%%%%%%%
\begin{abstract}
	With the advancement of robotics, machine learning, and machine
perception, increasingly more robots will enter human environments to
assist with daily tasks. However, dynamically-changing human
environments requires reactive motion plans. Reactivity can be
accomplished through re-planning, e.g. model-predictive control, or
through a reactive feedback policy that modifies on-going behavior in
response to sensory events. In this paper, we investigate how to use
machine learning to add reactivity to a previously learned nominal
skilled behavior. We approach this by learning a reactive modification
term for movement plans represented by nonlinear differential
equations. In particular, we use dynamic movement primitives (DMPs) to
represent a skill and a neural network to learn a reactive policy from
human demonstrations. We use the well explored domain of obstacle
avoidance for robot manipulation as a test bed.
Our approach demonstrates how a neural network can be combined with
physical insights to ensure robust behavior across different obstacle
settings and movement durations.
Evaluations on an anthropomorphic robotic system demonstrate the
effectiveness of our work.
\end{abstract}

%%%%%%%%%%%%%%%%%%%%%%%%%%%%%%%%%%%%%%%%%%%%%%%%%%%%%%%%%%%%%%%%%%%%%%%%%%%%%%%%

%
\section{Introduction}\label{sec:introduction}
\AR{Giovanni: DONE : We should shorten the introduction a little bit
  if we want to add Giovanni's figure. I don't think intro+abstract
  should take more than a page.}
\GS{Giovanni: DONE : So far this is
  the shortest version I could think of.}
In order to become effective assistants in natural human environments,
robots require a flexible motion planning and control approach. For
instance, a simple manipulation task of grasping an object involves a
sequence of motions such as moving to the object and grasping it.
While executing these plans, several scenarios can create the need to
modulate the movement online. Typical examples are reacting to changes
in the environment to avoid collisions, or adapting a grasp skill to
account for inaccuracies in object representation.

Dynamic movement primitives (DMPs) \cite{Ijspeert_NC_2013} are one
possible motion representation that can potentially be such a reactive
feedback controller. DMPs encode kinematic control policies as
differential equations with the goal as the attractor. A nonlinear
forcing term allows shaping the transient behavior to the attractor
without endangering the well-defined attractor properties. Once this
nonlinear term has been initialized, e.g., via imitation learning,
this movement representation allows for generalization with respect to
task parameters such as start, goal, and duration of the movement.

The possibility to add online modulation of a desired behavior is one
of the key characteristics of the differential equation formulation of
DMPs. This online modulation is achieved via coupling term functions
that create a forcing term based on sensory information -- thus
creating a reactive controller. The potential of adding feedback terms
to the DMP framework has already been shown in a variety of different
scenarios, such as modulation for obstacle avoidance
\cite{Park_Humanoids_2008,Hoffmann_ICRA_2009, Zadeh_AutonRob_2012,
  rai2014learning} and adapting to force and tactile sensor feedback
\cite{Chebotar_IROS_2014, Gams_TransRob_2014}. These approaches have
relied on extensive domain knowledge to design the form of the
feedback term. But we would like to realize all these behaviors within
one unified machine learning framework.
This goal opens up several problems such as, how to combine several
domain-specific coupling terms without extensive manual intervention and
how to design such a compact representation of the coupling term while
maintaining generalizability across varying task parameters.
% This goal opens up several
% problems like, how to combine several domain-specific coupling terms
% without extensive manual intervention, how to generalize coupling
% terms across varying task parameters, and how to design such a compact
% representation of the coupling term while maintaining
% generalizability.

%
\begin{figure}[t]
	\centering
	% \begin{subfigure}[t]{0.5\textwidth}
    \includegraphics[width=0.9\columnwidth]{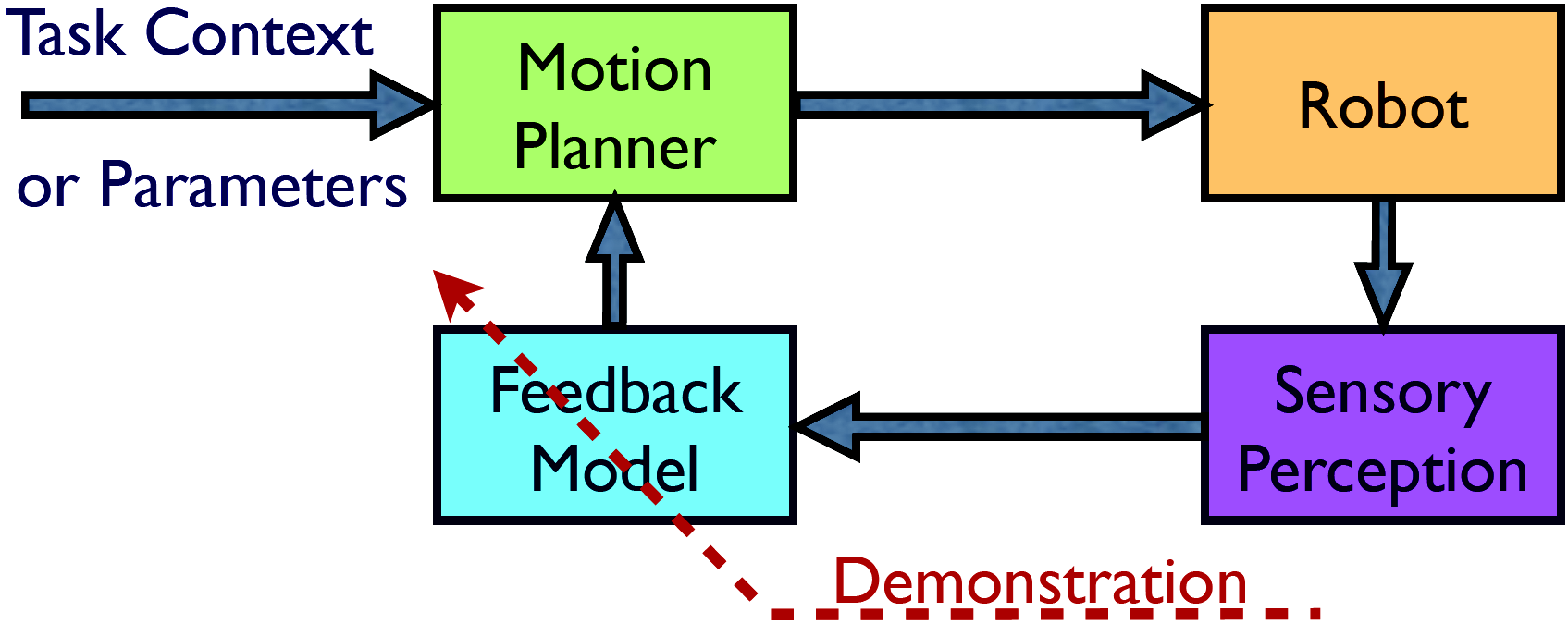}
 		\caption{Proposed framework for learning feedback terms.}
		\label{fig:BlockDiagramLearnFbTerms}
    \vspace{-0.3cm}
\end{figure}
In this paper, we investigate some first steps towards a more general
approach to learning coupling term functions. We present contributions
along two major axis: Part of our work is concerned with generalizing
DMPs with learned forcing and coupling terms. Towards this, we discuss
a principled method of creating a local coordinate system of a DMP and
creating duration invariant formulations of coupling terms. As a
result, demonstrations with different task parameters become
comparable. Additionally, we propose to choose a representation of feedback
terms that has the inherent potential to incorporate a variety of
sensory feedback. Similar to learning the shape of motion primitives -
we would like to be able to initialize such a general representation
using human demonstrations, to learn the mapping from sensory feedback
to coupling term. The overall system diagram is depicted in Figure
\ref{fig:BlockDiagramLearnFbTerms}. Given such a general coupling term
representation we then would like to incorporate some of the physical
intuition -- typically used to design the coupling term representation
-- to create robust and safe behaviors.
% \GS{adding tau-invariance:}
% \AR{I think this is a 4th contribution. Or we should combine this with
%   the local coordinate system, maybe?} Third, we show how we can turn
% our representation of feedback terms to become invariant with respect
% to motion duration.

% The long term vision would involve
% reinforcement learning to then refine the initialized feedback term
% representation.
This paper is organized as follows. We start out by reviewing
background on DMPs and the use of coupling terms in
Section~\ref{sec:background}. We then describe how we implement local
coordinate transformations within our system in
Section~\ref{sec:system_overview}. This is followed by the details of
our coupling term learning approach in
Section~\ref{sec:feedback_learning}. Finally, we evaluate our approach
in Section~\ref{sec:experiments} and conclude with
Section~\ref{sec:discussion}.

\section{Background} \label{sec:background}
\GS{Giovanni : Done: Define all the terms in the equations of DMP.}
We need a representation of planning and control for our work that
allows for a flexible insertion of machine learning terms to adapt
the planned behavior in response to sensory events.
% In
% \cite{Ijspeert_NC_2013}, a discussion of this topic is
% provided, and it is concluded that movement representations based on
% differential equations, often called ``pattern generators'', offer a
% favorable approach, as differential equations enable the insertion
% of additive coupling terms to modify behaviors in a reactive
% manner.
Dynamic Movement Primitives (DMPs) \cite{Ijspeert_NC_2013} are
one possibility of such representation, and we
adapt the DMP approach for our work due to its convenient and
well-established properties.

In brief, DMPs allow us to learn behaviors in terms of nonlinear
attractor landscapes. Integrating the DMP equations forward in time
creates kinematic trajectory plans, that are converted into motor
commands by traditional inverse kinematics and inverse dynamics
computations. The DMP differential equations have three components:
the main equation that creates the trajectory plan (called a
transformation system), a timing system (called canonical system), and
a nonlinear function approximation term to shape the attractor
landscape (called forcing term). Let $x$, $\xd$ and $\xdd$ represent
position, velocity and acceleration of the trajectory, then the
\textit{transformation system} can be written as follows:
\begin{equation} 
	\begin{split}
		\tau^2 \ddot{x} &= \alpha_v \left( \beta_v \left( g - x \right) - \tau \dot{x} \right) + a f  + C_t \\
%		\tau \dot{x} &= v
	\end{split}
        \label{eq:SchaalDMPTransformationSystem}
\end{equation}
for a one-dimensional system, where $\tau$ is the movement duration. 
The nonlinear forcing term $f$ is scaled by 
$a = \frac{g - x_{0}}{g_\text{demo} - x_{0,\text{demo}}}$, the ratio of distances between the start position $x_0$ and the goal position $g$ during unrolling and during demonstration.
%
% The parameters are usually set $\alpha_v = 25$ and
% $\beta_v = \alpha_v/4$ such that the transformation system without $f$
% and coupling term $C_t$ becomes a critically-damped second order
% dynamical system that converges to goal $g$.
%
The \textit{canonical system} defines phase variable $s$, representing
the current phase of the primitive. This component of the DMP adds the
ability to scale a motion primitive to different durations. The
canonical system is a first-order dynamical system, given by
\begin{equation} 
	\tau \dot{s} = -\alpha_s s
	\label{eq:2ndOrderCanonicalSystem}
\end{equation}
%

% The first order parameter is usually chosen $\alpha_s = 25/3$. In both
% tranformation system and canonical system, $\tau$ is the parameter
% that defines the motion duration.
%
%
The transformation system is driven by a nonlinear \textit{forcing
  term} $f$ and a coupling term $C_t$. The forcing term $f$ creates
the nominal shape of a primitive and is typically modeled as a
weighted sum of $N$ Gaussian basis functions $\psi_i$ which are
functions of the phase $s$, with width parameter $h_i$ and center at
$c_i$, as follows:
\begin{equation} 
  f\left( s \right) = \frac{\sum_{i=1}^N \psi_i \left( s \right) w_i}{\sum_{i=1}^N \psi_i \left( s \right)} s
  \label{eq:DMPForcingTerm2ndOrderCanonicalSys}
\end{equation}
where
%The forcing term is a weighted sum of basis functions:
\begin{equation} 
  \psi_i \left( s \right) =  \exp\left( -h_i \left( s - c_i \right)^2 \right)
  \label{eq:GaussianBasisFunction}
\end{equation}
The forcing term weights $w_i$ are learned from human demonstration,
as pointed out in \cite{Ijspeert_NC_2013}. The influence of $f$
vanishes as $s$ decays to 0, and as a result, position $x$ converges
to the goal at the end of the movement.
% The evolving goal state $g$ is the smoothed version of $G$, which
% ensures there will be no discontinuity in the transformation system
% when the goal position $G$ is changed.
Beside the forcing term, the transformation system could also be
modified by the \textit{coupling term} $C_t$, a sensory coupling,
which can be either state-dependent or phase-dependent or both.
% \GS{adding tau-invariance:} \AR{I don't think the next sentence about
%   tau-invariance is needed in the background section. But I won't take
%   it out.} Interestingly, the inclusion of $\tau$ as a component in
% the feature set for coupling term $C_t$, as will be pointed out in
% section \ref{sec:experiments} enables a coupling term that is
% invariant with respect to motion duration ($\tau$-invariant).

% At any moment of time, $x$, $\xd$ and $\xdd$ represent the desired
% state of a DMP, which is given as a trajectory plan to the robot
% controller.
%
For a multi degree-of-freedom (DOF) system, each DOF
has its own transformation system, but all DOFs share the same
canonical system \cite{Ijspeert_NC_2013}.

\subsection{Coupling Terms}\label{sec:previous_coupling_term}
\AR{Akshara : Done: Shorten the coupling term work.}
The coupling term $C_t$ in Equation
\ref{eq:SchaalDMPTransformationSystem} plays a significant role in
this paper and deserves some more discussion.
%In the same way as the forcing term $f$, the 
Coupling terms can be used to modify a DMP on-line, based on any state
variable of the robot and/or environment. Ideally, a coupling term
would be zero unless a special sensory event requires to modify the
DMP. One could imagine a coupling term library that handles a variety
of situations that require reactive behaviors. In the past, coupling
terms have been used to avoid obstacles \cite{rai2014learning}, to
avoid joint-limits \cite{gams2009line} and to grasp under uncertainty
\cite{Pastor_ICRA_2009}. Coupling terms from previous executions can
also be used to associate sensory information with the task, as
proposed in \cite{pastor2013dynamic}.

Obstacle avoidance is a classical topic in the motion planning
literature. In reference to DMPs, several papers have tried to develop
coupling term models that can locally modify the planned DMP to avoid
obstacles. Park et al. \cite{Park_Humanoids_2008} used a dynamic
potential field model to derive a coupling term for obstacle
avoidance. Hoffman et al. \cite{Hoffmann_ICRA_2009} used a
human-inspired model for obstacle avoidance, and Zadeh et al.
\cite{Zadeh_AutonRob_2012} designed a multiplicative (instead of
additive) coupling term. Gams et al. \cite{gams2014learning} directly
modify the forcing term $f$ of a DMP in an iterative manner and apply
it to the task of wiping a surface.
% through an external input, which is a control law that provides a force reference to the robot. 
This is a step towards automatically learning coupling terms based on
experience, rather than hand-designed and hand-tuned models. Chebotar
et al. \cite{Chebotar_IROS_2014} also used reinforcement learning to
learn a tactile-sensing coupling term, modulated by tactile feedback
from the sensors.
%Parameters are learned in an iterative manner using episodic relative entropy policy search. 
More recently, Gams el al. expanded
their work in \cite{gams2015learning}, by generating
a database of coupling terms and generalizing to multiple scenarios. 

%In case an existing coupling term is not found in the database, it is learnt iteratively using Iterative Learning Conrol.
All of the above approaches take an iterative approach towards
learning the parameters of their coupling term model, but suffer from a lack of
generalizability to unseen settings. While any new setting can be
learned afresh, there is useful information in every task performed by
a robot that can be transferred to other tasks.

\begin{figure}[t]
	\centering
	\includegraphics[width=0.35\textwidth]{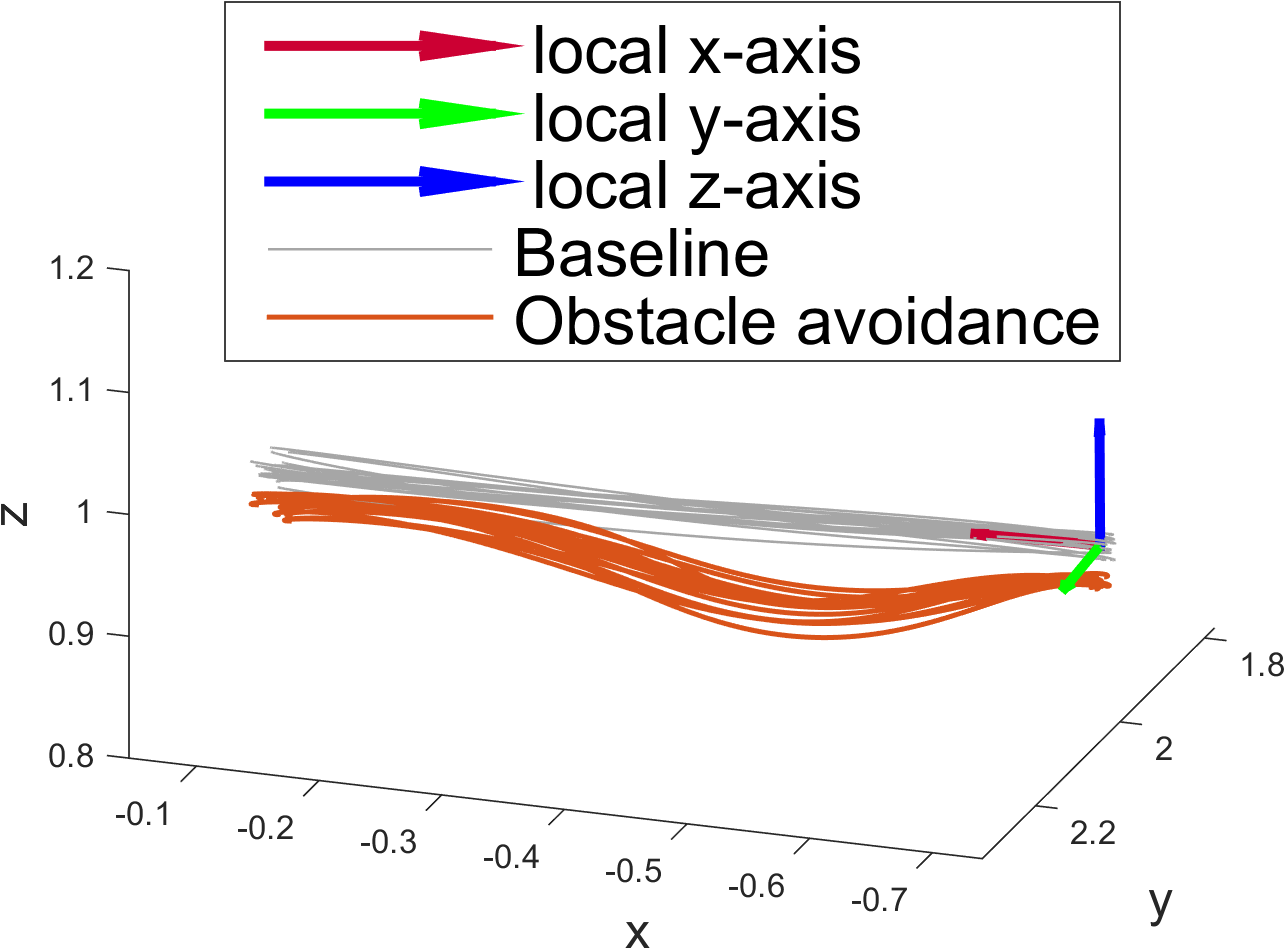}
	\\
	\includegraphics[width=0.23\textwidth]{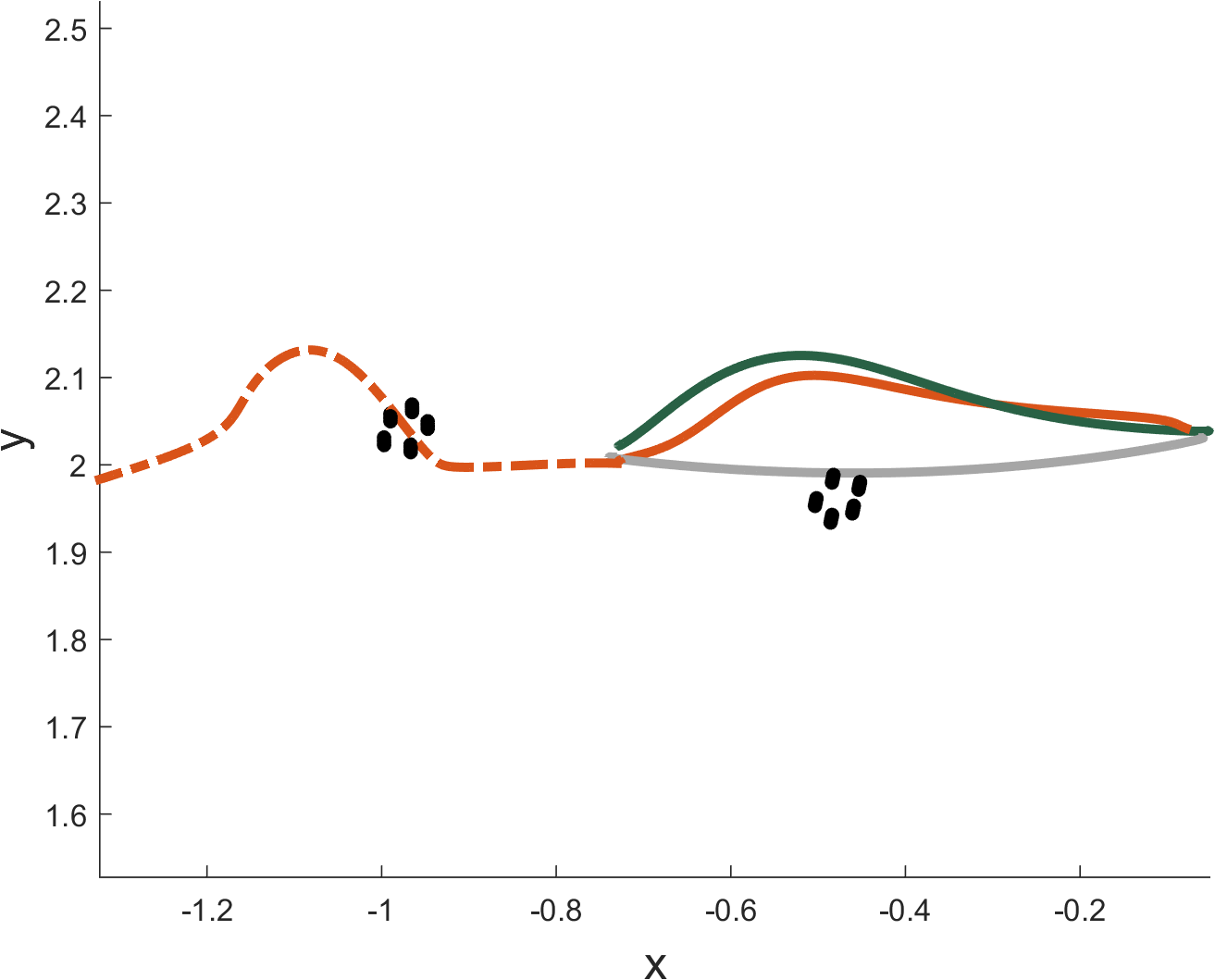}
	\includegraphics[width=0.23\textwidth]{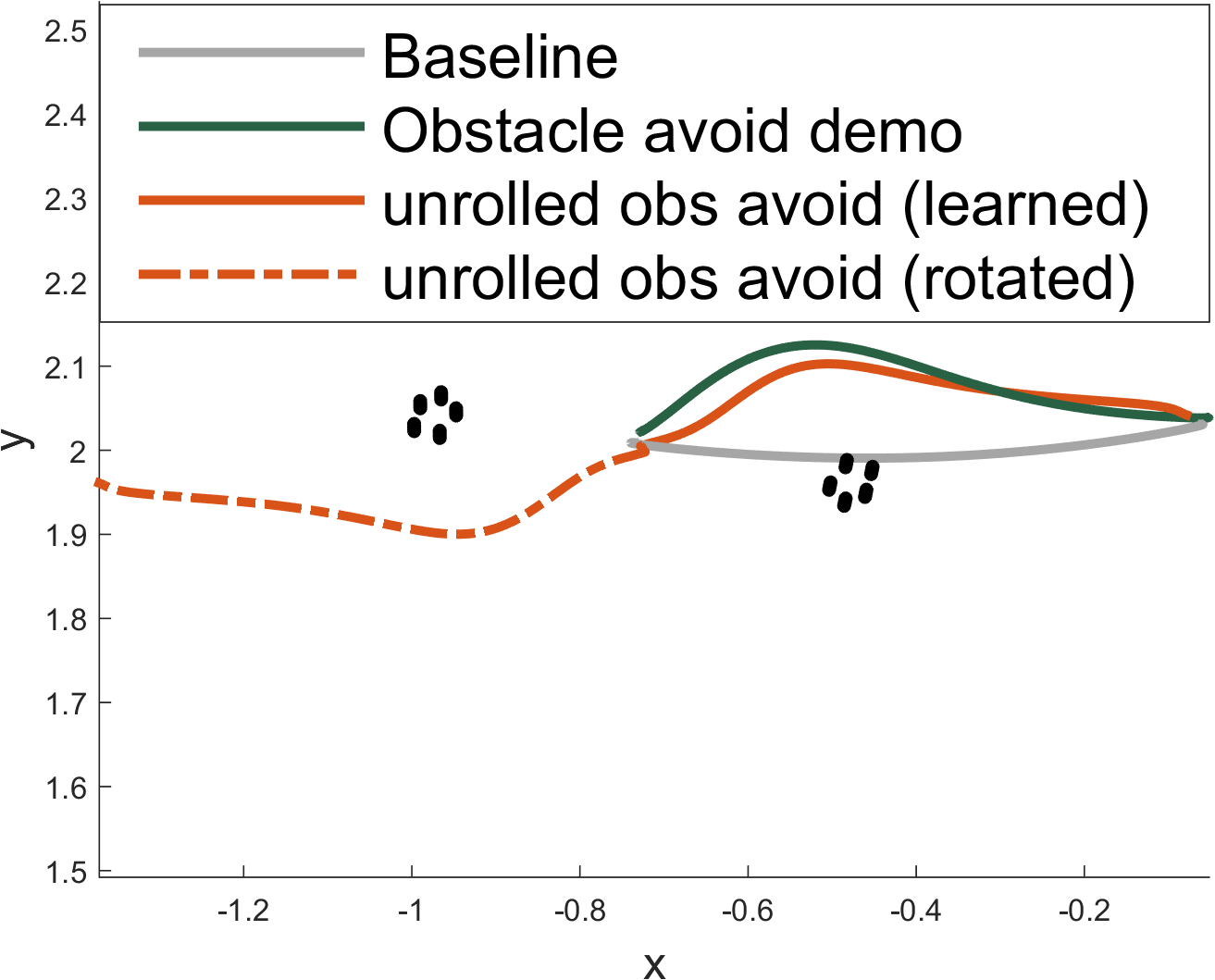}%
	\caption{(top) Example of local coordinate frame definition
          for a set of obstacle avoidance demonstrations. A local coordinate frame is defined on trajectories collected from human demonstration. (bottom) Unrolled avoidance behavior is shown 
          for two different location of the obstacle and the goal:
          using local coordinate system definition (bottom right) 
          and not using it (bottom left).}
%          Using local coordinates, the bottom right plot demonstrates
%          a more reasonable obstacle avoidance behavior than the bottom left plot.}
	\label{fig:LocalCoordinateSystemExample}
\end{figure}

Hand-designed features can extract useful information from the
environment, but it can be hard to find and tune such hand-designed
features. In our previous work \cite{rai2014learning}, we tried to
start with human-inspired features for coupling terms from
\cite{Hoffmann_ICRA_2009} and learn parameters for these features
using human demonstrations.
%To generalize across multiple obstacles, we initialized the features on a linear grid and used Automatic Relevance Determination (ARD) \cite{li2002bayesian} to get rid of redundant features.
This model could generate human-like obstacle avoidance movements for
one setting of demonstrations for spherical and cylindrical obstacles.
However, it did not generalize across different obstacle avoidance
settings.

%using ARD makes the learned weights rather high in magnitude and it was necessary to stop the process of eliminating features prematurely.
%Also, the authors mention that their model does not generalize well across different obstacle avoidance settings.

\GS{Akshara: the following paragraph is similar to those in the introduction section, 2nd to last part. Either this one or the one in the intro needs to be eliminated to shorten the paper.}
\AR{I've taken out the lengthy contribution, and just kept what's relevant here now.}

In this paper, we propose a neural-network based coupling model. Given
human data this model can be trained to avoid obstacles, and
generalizes to multiple obstacle avoidance settings. This eliminates
the need for hand-designed features, as well as results in robust
obstacle avoidance behavior in unseen settings.
%  
%Thus, our contribution in this paper is manyfold. First, we define a local coordinate system for DMPs that allows generalizing learned coupling terms across multiple obstacle scenarios. Second, we propose a neural-network based obstacle avoidance model that can fit human data offline and generalize to different obstacle avoidance settings, eliminating hand-designed features. Third contribution is making this neural network safe for obstacle avoidance such that we can be sure that the neural network outputs do not make our robot unstable in any unseen situation. 
%\GS{adding tau-invariance:} Fourth, we show how we can turn our representation of feedback terms to become invariant with respect to motion duration.

%
%
%
\section{Spatial Generalization using Local Coordinate Frames}

\AR{Akshara : Done : Change figure and description. Unit length vectors make the figure look ugly, as they are too long. I don't think the figure has to have unit vectors, the global coordinates are not unit vectors either. I think the description of the local coordinates should say unit vectors, which Giovanni fixed.}
\GS{Giovanni : Scale is wrong again, it is NOT unit length now, please fix it.}
%
% \subsection{Local Coordinate Transform}\label{sec:local_coord_transform}
%
% 
%Ijspeert et al. \cite{Ijspeert_NC_2013} pointed out the importance of
%the definition of coordinate systems for multi-dimensional DMPs. For
%instance, choosing a 3D task space as fixed Cartesian coordinate
%system, as a cylindrical coordinate system, or as a spherical
%coordinate system can have a profound impact in the qualitative
%behavior of how a DMP generalizes to new goals.  Thus, in order to
%learn reactive coupling terms for obstacle avoidance, we need to
%ensure that these coupling terms create consistent behavior
%independent of the start and goal of a DMP.
%
%Following an idea of using a radial coordinate system for two-dimensional
%DMPs in Ijspeert et al. \cite{Ijspeert_NC_2013}, we define a local
%coordinate system for a three-dimensional task space DMP as follows:
Ijspeert et al. \cite{Ijspeert_NC_2013} pointed out the importance of a
local coordinate system definition for the spatial-generalization of
two-dimensional DMPs. Based on this, we define a three-dimensional
task space DMPs as follows:
\begin{enumerate}
\item Local x-axis is the unit vector pointing from the start position towards the goal position.
\item Local z-axis is the unit vector orthogonal to the local x-axis
  and closest to the opposite direction of gravity vector.
\item Local y-axis is the unit vector orthogonal to both local x-axis
  and local z-axis, following the right-hand convention.
\end{enumerate}
% this is a sentence for a conclusion
% We found that this coordinate system definition is particularly
% important in spatially generalizing the learned reactive obstacle
% avoidance coupling terms across different start and goal positions.

The first figure on the top of Figure
\ref{fig:LocalCoordinateSystemExample} gives an example of a local
coordinate system defined for a set of human obstacle avoidance
demonstrations.
  
\begin{figure}[t]
	\centering
	% \begin{subfigure}[t]{0.5\textwidth}
   \includegraphics[width=0.9\columnwidth]{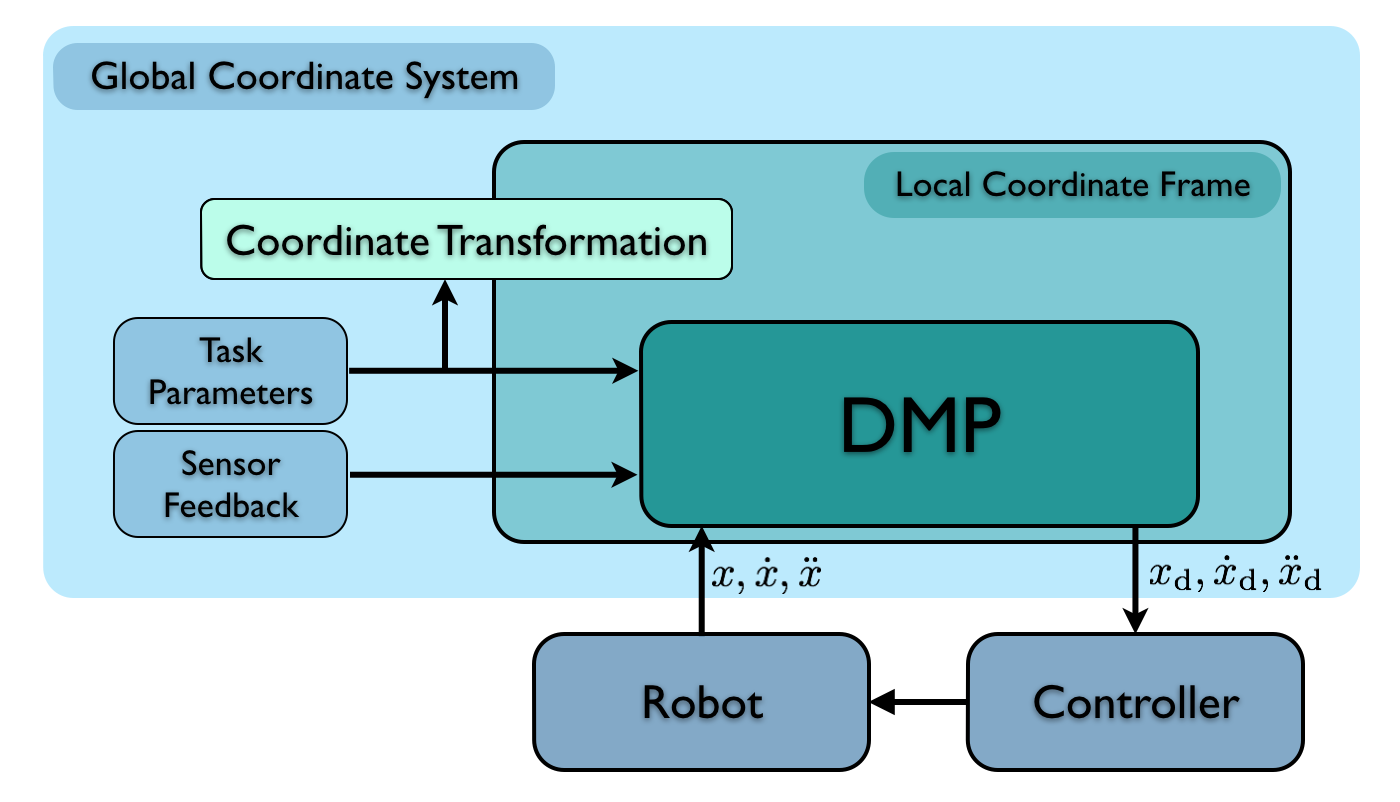}
		\caption{System overview with local coordinate transform.}
		\label{fig:BlockDiagramLearnObsAvoid}
\end{figure}

The importance of using a local coordinate systems for obstacle
avoidance is illustrated in Figure
\ref{fig:LocalCoordinateSystemExample} bottom plots. In both plots,
black dots represent points on the obstacles. Solid orange trajectories represent
the unrolled trajectory of the DMP with learned coupling term when the
goal position is the same as the demonstration (dark green). Dotted orange
trajectories represent the unrolled trajectory when both goal position
and the obstacles are rotated by 180 degrees with respect to the start. DMPs without local coordinate system (bottom left) are
unable to generalize the learned coupling term to this new task
setting, while DMPs with local coordinate system (bottom right) are
able to generalize to the new context. When using local coordinate
system, all related variables are transformed into the representation
in the local coordinate system before using them as features to
compute the coupling term, as described in Figure
\ref{fig:BlockDiagramLearnObsAvoid}.

\AR{I think this figure is important if someone wants to implement the local coordinate system.}

\label{sec:system_overview}

%  \begin{figure}[h]
%  	\centering
%  	\begin{subfigure}[t]{0.5\textwidth}
%  		\includegraphics[trim={0.5cm 0.9cm 0.5cm 1.1cm},clip,width=\textwidth]{figures/local_coord_system/not_using_local_coord_transform.png}
%  		\caption{Not using local coordinate frame.}
%  		\label{fig:NotUsingLocalFrame}
%  	\end{subfigure}
%      \begin{subfigure}[b]{0.5\textwidth}
%  		\includegraphics[trim={0.5cm 0.9cm 0.5cm 1.1cm},clip,width=\textwidth]{figures/local_coord_system/using_local_coord_transform.png}
%  		\caption{Using local coordinate frame.}
%  		\label{fig:UsingLocalFrame}
%  	\end{subfigure}
%  	\caption{Generalization of task context in obstacle avoidance: example of unrolled trajectories from a learned obstacle avoidance model under rotation in z-axis.}
%  	\label{fig:LocalFrameTest}
%  \end{figure}

%
\section{Towards General Feedback Term Learning}
\label{sec:feedback_learning}
%
% A general coupling term approach entails automatic learning features
% from data, and generalizing to unseen situations. While there have
% been automatically learnt features for coupling terms, as well as
% hand-designed features that generalize to unseen situations, there is
% no prior work on learning features that are also capable at
% generalizing to new scenarios.

% In previous work \cite{rai2014learning} we attempted to
% hand-design a coupling term model, but found this to not be flexible
% enough to fit human data across multiple demonstrations.

% In the future, we envision using coupling terms for objectives other
% than just obstacle avoidance - for example kinematic and dynamic
% constrains of a robot, using feedback for tracking, grasping, etc.
% Designing features for all these tasks can be cumbersome, and might
% not be powerful enough to fit over complicated datasets. Hence, we
% would like to create a general coupling term learning framework that
% can be used to automatically learn features that help complete a
% particular task. With this vision, we want to move towards a more
% general obstacle avoidance framework that is able to fit across
% multiple obstacle avoidance settings as well as generalize to unseen
% settings.

The larger vision of our work is to create a coupling term learning
framework that has the flexibility to incorporate various sensor
feedback signals, can be initialized from human data and can
generalize to previously unseen scenarios. We envision
using coupling terms for objectives other than just obstacle avoidance
- for example kinematic and dynamic constraints of a robot, using
feedback for tracking, grasping, etc. Towards this goal we present our
approach to general feedback term learning in the context of obstacle
avoidance.

One step towards generalizing to unseen settings is to use a
transformed coordinate system, as introduced in Section
\ref{sec:system_overview}. The second challenge of creating a flexible
coupling term model is addressed by choosing an appropriate function
approximator, that can be fit to predict coupling terms given sensory
feedback. Here, we choose to model the coupling term function through
a neural network which is trained on human demonstrations of obstacle
avoidance.
\AR{Added other neural network based work. Please edit/add
  more references, and change the write-up.}
Neural networks have been successfully applied in many different
applications -- including robotics -- and are our function
approximator of choice. Typically in robotics, neural networks are
used to directly learn the control policy in a model-free way, for
example in \cite{lillicrap,levine,pinto,zhang2015towards,gu2016deep}.
In these papers, deep networks directly process the visual input and
produce a control output.
%
% in a reinforcement learning fashion, like policy search
% or by learning a Q-function.
%
These approaches use reinforcement learning to learn policies from
scratch, or start with locally optimal policies or demonstrations.
This results in a very general learning control formulation that, in
theory, can generalize to almost any robot or task at hand.
%Some initial results  
% been successfully demonstrated on robots, for example in
% \cite{zhang2015towards}, and \cite{gu2016deep}.
In contrast to the
common model-free way, we would like to inject structure in our
learning through DMPs and use the neural network to locally modulate a
global plan created from a trajectory optimizer, or demonstration. We
expect such a structure to enable our control to scale to higher
dimensions, as well as generalize across different tasks.

While there is no question that neural networks have the necessary
flexibility to represent a coupling term model with various sensor
inputs, there is concern regarding their unconstrained use in
real-time control settings. It is likely that the system encounters
scenarios that have not been explicitly trained for, for which it is
not always clear what a neural network will predict. However, we want
to ensure that our network behaves safely in unseen settings. Thus, as
part of our proposed approach, we introduce some physically inspired
post processing measures that we apply to our network predictions
which ensure safe behaviors including convergence of the motion
primitive.

\subsection{Setting up the learning problem}
To learn a general coupling term model from human demonstrations we
follow a similar procedure as described in \cite{rai2014learning}. We
start by recording human demonstrations of point-to-point movements,
with and without an obstacle on different obstacle settings. The
demonstrations without obstacle are used to learn the forcing term
function $\hat{f}(s)$ of the basic dynamic movement primitive
representation. All demonstrations with obstacle avoidance behavior
are then used to capture the coupling term value with respect to the
assumed underlying primitive. For clarity purposes, we refer to the
primitive without obstacle avoidance as the baseline to make a
distinction from the motion primitive with obstacle avoidance.

% Each
% setting in our experiments has a unique obstacle position or shape. We
% collect several demonstrations of obstacle avoidance per settings.

% In our experiments, we keep the setting without obstacle constant
% across all obstacle avoidance demonstrations, and call it the baseline
% setting. This is done primarily to simplify the data collection as
% each obstacle avoidance and baseline movement make a pair. Instead of
% $2N$ settings, we collect data for $N+1$ settings and keep the
% baseline the same.
%
% Additionaly, our learning problem formulation tries
% to make the coupling term independent of the baseline.

The coupling term $C_t$ of a given demonstration can
be computed as the difference of forcing terms between obstacle
avoidance behavior and the baseline motion primitive. 
For a particular obstacle avoidance trajectory, this becomes
\begin{equation} \label{eq:new_Ct_target_computation}
	C_t = \tau^2 \ddot{x}_o - \alpha_v \left( \beta_v \left( g - x_o \right) - \tau \dot{x}_o \right) - a \hat{f}(s)
\end{equation}
where $x_o$, $\dot{x}_o$ and $\ddot{x}_o$ are the position, velocity
and acceleration of the obstacle avoidance trajectory. Since the start and goal
positions of the baseline and obstacle avoidance demonstrations are
the same in our training demonstrations, $a = 1$ for the fitting
process. Furthermore, $\tau$ is the movement duration and $\alpha_v$
and $\beta_v$ are constants defined in Section \ref{sec:background}.

By computing the difference in forcing terms between the baseline
primitive and the obstacle avoidance demonstration, we capture the
quantity $C_t$ that our coupling term model should essentially
predict. Further, this formulation makes the target coupling term
relatively independent of the baseline trajectory and can also easily
handles different lengths of trajectories.

The target coupling term $C_t$ is calculated for all the
demonstrations and concatenated, giving us the regression target
$\mathbf{C_{t}}$. Our goal now is to learn a function $h$, mapping
sensory features $\mathbf{X}$ -- extracted from the demonstrations --
to targets $\mathbf{C_t}$:
% This $\mathbf{C_{t}}$ is a function of the obstacle
% avoidance trajectories and we want to learn a representation between
% trajectory features $\mathbf{X}$ and the target coupling term.
% \AR{confusing with the forcing term?}
\begin{equation}
\mathbf{C_{t}} = h(\mathbf{X})
\end{equation}
This is a general regression problem which can be addressed using any
non-linear function approximator.
% Next we desribe how we use neural network 
%
\subsection{Coupling Term Learning with Neural Networks}
Neural networks are powerful non-linear function approximators that
can be fast and easy to deploy at test time. Given their
representational power, neural networks seem to fit into our larger
vision of this work. Generally speaking however, any non-linear
function approximator could be considered for this part of the
framework.
% Given their implementation simplicity, we decided
% to use neural networks for our problem, though it could have been
% solved with other non-linear function approximators too.

Here, the target coupling term is approximated as the output of our neural
network, given sensory features of the obstacle avoidance demonstration.
\begin{equation}
\mathbf{C_{t}} = h_\text{NN}(\mathbf{X})
\end{equation} 
The inputs $\mathbf{X}$ are extracted from the obstacle avoidance
demonstration. Details of the components of the feature vector $\mathbf{X}$ 
are explained in Section \ref{sec:experiments}.
%We consider the following inputs for our network:
%\begin{enumerate}
%\item vector between points on the obstacle and end-effector 
%\item vector between obstacle center and end-effector
%\item motion duration ($\tau$)-scaled velocity of end-effector ($\tau v$)
%\item distance to the obstacle
%\item angle between the end-effector velocity and obstacle
%\end{enumerate}
%
%These inputs are illustrated in Figure \ref{obs_avoid_3D}.
%\begin{figure}
%\centering
%\includegraphics[scale = 0.5]{figures/obs_avoid_3D.png}
%\caption{Illustration of the inputs to the network}
%\label{obs_avoid_3D}
%\end{figure}

Since we consider meaningful input features - that we believe to have
an influence on obstacle avoidance behaviors - we do not require the
neural network to learn this abstraction, although this would be an
interesting avenue for future work. Because of this we only require a
shallow neural network, with three small layers only. The hidden
layers have rectified linear units (ReLU \cite{nair2010rectified}) and
the output layer is a sigmoid, such that the output is bounded. We
train one neural network on the three-dimensional target coupling
term. Weights and biases are randomly initialized and trained using
the Levenberg-Marquardt algorithm. We use the MATLAB Neural Networks
toolbox in our experiments \cite{MATLAB:2015a}.

% this should go in experimetnal section
% This gives us a total of 17 inputs to the network. The neural network
% structure has a depth of 3 layers, with 2 hidden layers with 20 and 10
% units each and an output layer. 
\subsection{Post-processing the neural network output}
\AR{ Done: Post-processing for multiple obstacles. I think I motivated
  the post-processing fine now. But please change it, if you can think
  of something. I would like to address the concern about how this
  post-processing is obstacle avoidance specific in the discussion.}
Particular care has to be taken when applying neural network
predictions in a control loop on a real system. Extrapolation behavior
for neural networks can be difficult to predict and comes without any
guarantees of reasonable bounds in unseen situations. In a problem
like ours, it is nearly impossible to collect data for all possible
situations that might be encountered by the robot. As a result, it is
important to apply some extra constraints, based on intuition, on the
predictions of the neural network. The final coupling term $C_t$,
given a set of inputs $x$ becomes
\begin{equation}
C_t = P(h_{NN}(x))
\end{equation}
where $P$ are the post-processing steps applied to the network's output
to ensure safe behavior. 

One common problem is that in some situations, we physically expect
the coupling term to be 0 or near 0. But due to noise in human data,
$\mathbf{C_t}$ is not necessarily 0 in these cases. For instance,
after having avoided the obstacle, we should ensure goal convergence
by preventing the coupling term from being active.
% However, Due to its
% flexibility, the neural network can learn to output a non-zero value
% in such cases.
With such cases in mind, the external constraints applied to
the output of the neural network while unrolling are as follows:

\begin{enumerate}
\item \textbf{Set coupling term in x-direction as 0: }In the
  transformed local coordinate system, the movement of the obstacle
  avoidance and the baseline trajectory are identical in the
  x-direction. This means that the coupling term in this dimension can
  be set to 0.
  The post-processed coupling term becomes
  \begin{equation}
  P((C_{tx}, C_{ty}, C_{tz})) = (0, C_{ty}, C_{tz})
  \end{equation}
\item \textbf{Exponentially reduce coupling term to 0 on passing the
    obstacle: } We would like to stop the coupling term once the robot
  has passed the obstacle, to ensure convergence to the goal. In the
  local coordinate frame, this can be easily realized by comparing the
  x-coordinate of the end effector with the obstacle location. To
  adjust to the size of the obstacle and multiple obstacles, this
  post-processing can be modified to take into account obstacle size
  and the location of the last obstacle. We exponentially reduce the
  coupling term output in all dimensions once we have passed the
  obstacle. The post-processing becomes:
  \[
  P(C_t) = 
  \begin{cases}
  C_t \; \text{exp}^{(-(x_o - x_{ee})^2)} , &\text{if } x_o < x_{ee} \\
  C_t, &\text{otherwise}
  \end{cases}
  \]
  where $x_o$ is the x-coordinate of the obstacle and $x_{ee}$ is the
  x-coordinate of the end-effector.
\item \textbf{Set coupling term to 0 if obstacle is beyond the goal: }
  If the obstacle is beyond the goal, the coupling term should
  technically be 0 (as humans do not deviate from the original
  trajectory). This is easily taken care of by setting the coupling
  term to 0 in such situations.
 \[
  P(C_t) = 
  \begin{cases}
  (0,0,0), &\text{if } x_o > x_{goal} \\
  C_t, &\text{otherwise}
  \end{cases}
  \]
  where $x_o$ and $x_{goal}$ are the x-coordinates of obstacle and
  goal respectively.
\end{enumerate}
Note, how all the post-processing steps leverage the local coordinate
transformation. This post-processing, while not necessarily helping
the network generalize to unseen situations, makes it safe for
deploying on a real robot. With this learning framework, and the local
coordinate transformation we are now ready to tackle the problem of
obstacle avoidance using coupling terms. In the next section, we
describe our experiments that use this framework to learn a network
and then deploy it as a feedback term in the baseline DMP.

\section{Experiments} \label{sec:experiments}
We evaluate our approach in simulation and on a real system. 
First, we use obstacle avoidance demonstrations 
collected as detailed below, to extensively evaluate
our learning approach in simulation. In the simulated obstacle
avoidance setting, we first learn a coupling term model and then
unroll the primitive with the learned neural network. We perform three
types of experiments: learning/unrolling per single obstacle setting,
learning/unrolling across multiple settings and unrolling on unseen
settings after learning across multiple settings. We also compare our
neural network against the features developed in
\cite{rai2014learning}. This involves defining a grid of hand-designed
features and using Bayesian regression with automatic relevance
determination to remove the redundant features. We are using three
performance metrics to measure the performance of our learning
algorithm:
\begin{enumerate}
\item Training NMSE (normalized mean squared error), calculated as
  the mean squared error between target and fitted coupling term,
  normalized by the variance of the regression target:
  \begin{equation}
   \text{NMSE} = \frac{\frac{1}{N}\sum_{n=1}^N\left(C_{t,n}^\text{target} - C_{t,n}^\text{fit}\right)^2}{\text{var}(C_{t}^\text{target})}.%
  \end{equation}
  where $N$ is the number of data points.
\item Test NMSE on a set of examples held out from the training.
\item Closest distance to the obstacle of the obstacle avoidance trajectory.
\item Convergence to the goal of the obstacle avoidance trajectory.
\end{enumerate}
Finally, we train a neural network across multiple settings and deploy
it on a real system.

In all our experiments detailed below we use the same neural network
structure: The neural network has a depth of 3 layers, with 2 hidden
layers with 20 and 10 ReLU units each and an output sigmoid layer. The
total number of inputs is 17 and the number of outputs is 3 for the
three dimensions of the coupling term.
\begin{enumerate}
\item vector between 3 points on the obstacle and end-effector (9 inputs)
\item vector between obstacle center and end-effector (3 inputs)
\item motion duration ($\tau$)-multiplied velocity of end-effector
  ($\tau \mathbf{v}$, 3 inputs)
\item distance to the obstacle (1 input)
\item angle between the end-effector velocity and obstacle (1 input)
\end{enumerate}
\subsection{Experimental Setup}\label{sec:experimental_setup}
\begin{figure}[h!]
    \centering
    \begin{subfigure}{0.23\textwidth}
        \includegraphics[width=\textwidth]{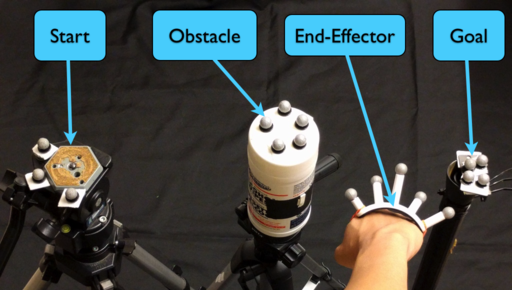}
        \caption{Data collection setting using Vicon objects to represent end-effector, obstacle,
          start and goal positions.}
        \label{subfig:demo_vicon_setting}
    \end{subfigure}
    \begin{subfigure}{0.23\textwidth}
        \includegraphics[width=\textwidth]{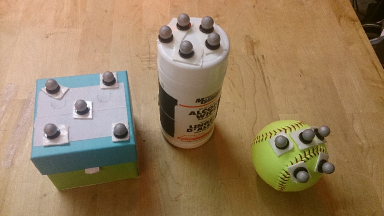}
        \caption{Different types of obstacles used in data collection,
          from left to right: cube, cylinder, and sphere}
        \label{subfig:obstacles}
    \end{subfigure}
    \caption{Data collection setting and different obstacle geometries
      used in experiment.}
    \label{fig:data_collection_setting}
\end{figure}

\begin{figure}[h]
%\begin{figure}[h!]
    \centering
    \begin{subfigure}[b]{0.235\textwidth}
      \includegraphics[trim={9cm 2cm 9cm 2cm},clip,width=\textwidth]{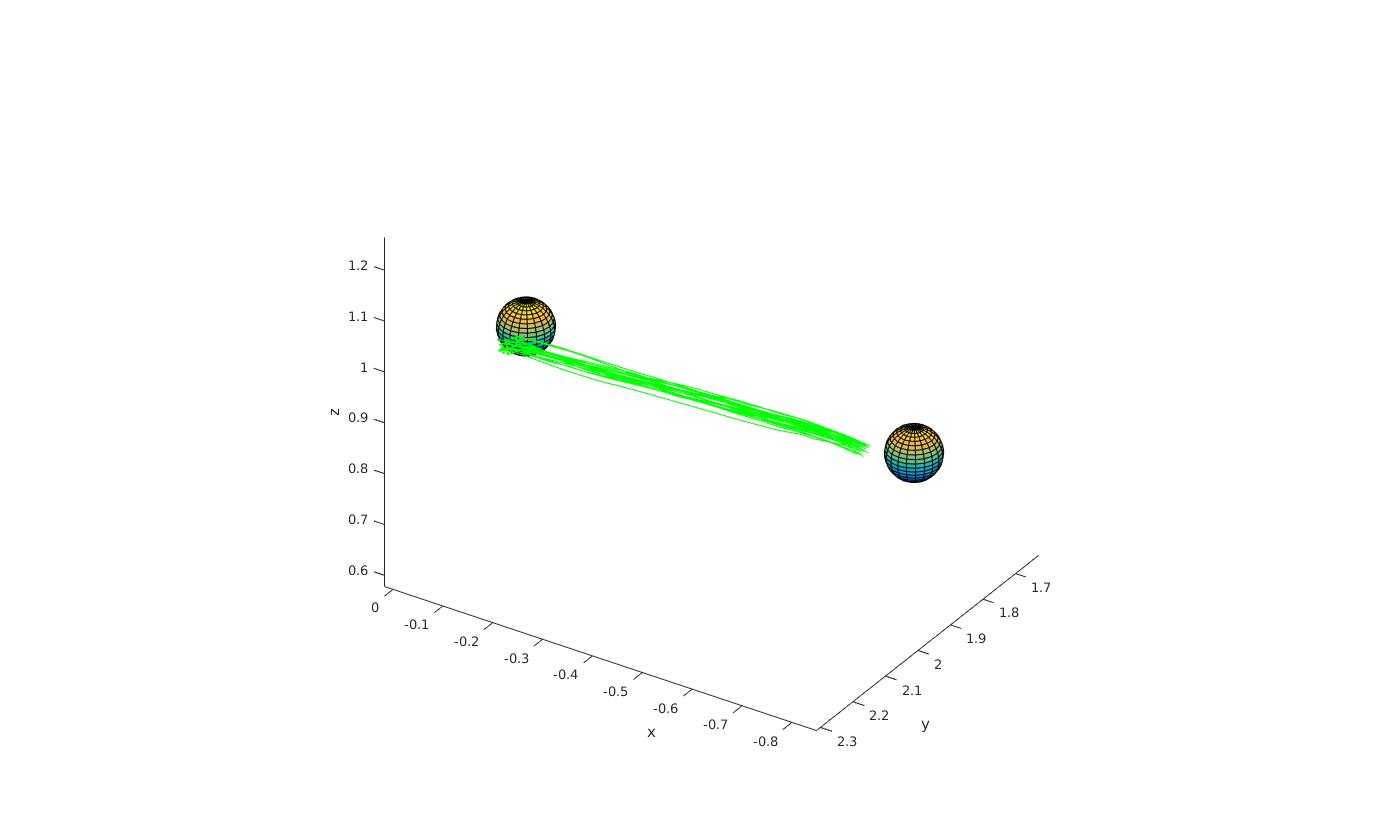}%
        \caption{All nominal/baseline demonstrations (no obstacles).}
        \label{subfig:baseline_all_good_samples}
    \end{subfigure}
    \begin{subfigure}[b]{0.235\textwidth}
        \includegraphics[trim={9cm 2cm 9cm 2cm},clip,width=\textwidth]{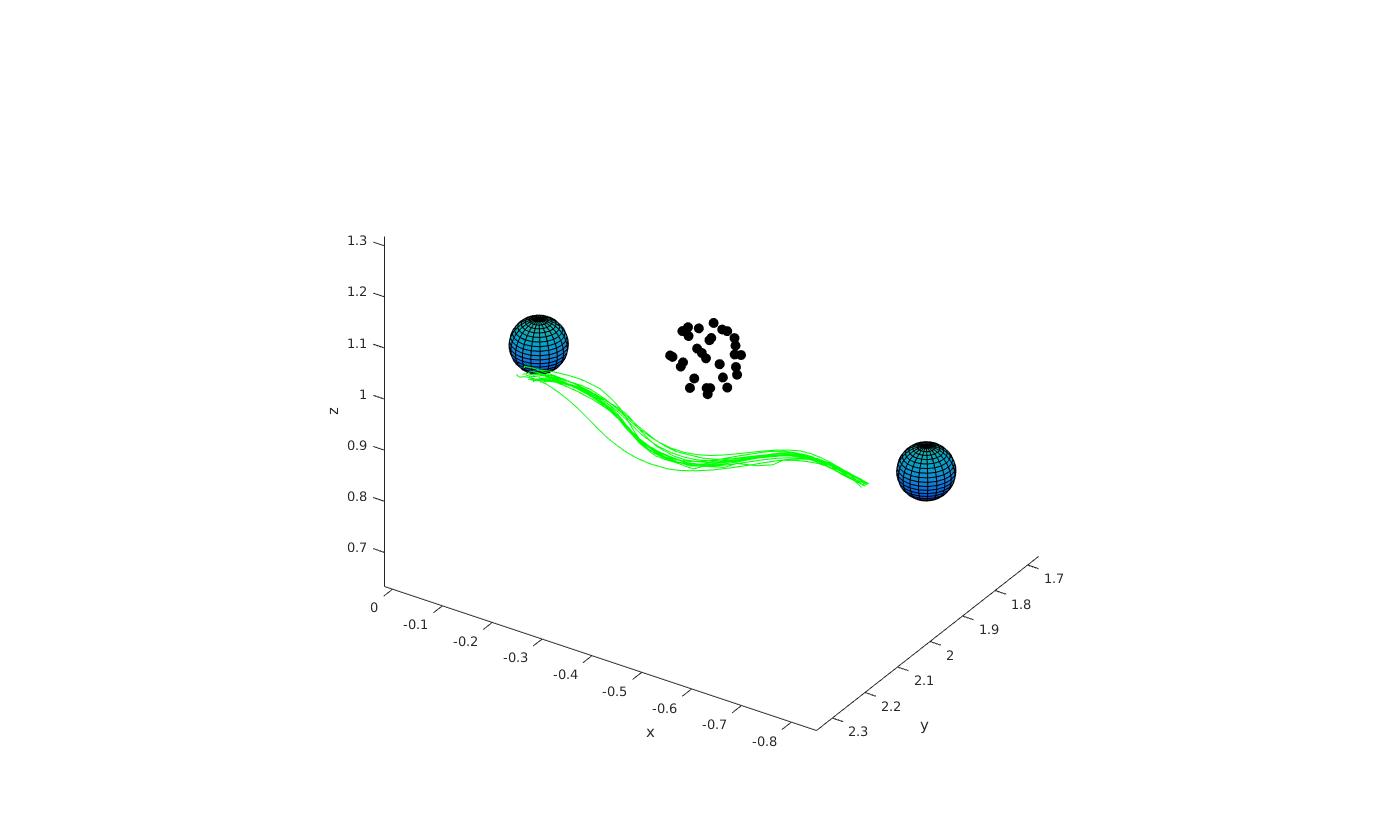}%
        \caption{Sphere obstacle avoidance demonstrations.}
        \label{subfig:sphere_all_good_samples_obs_avoid_27}
    \end{subfigure}
    \begin{subfigure}[b]{0.235\textwidth}
        \includegraphics[trim={9cm 2cm 9cm 2cm},clip,width=\textwidth]{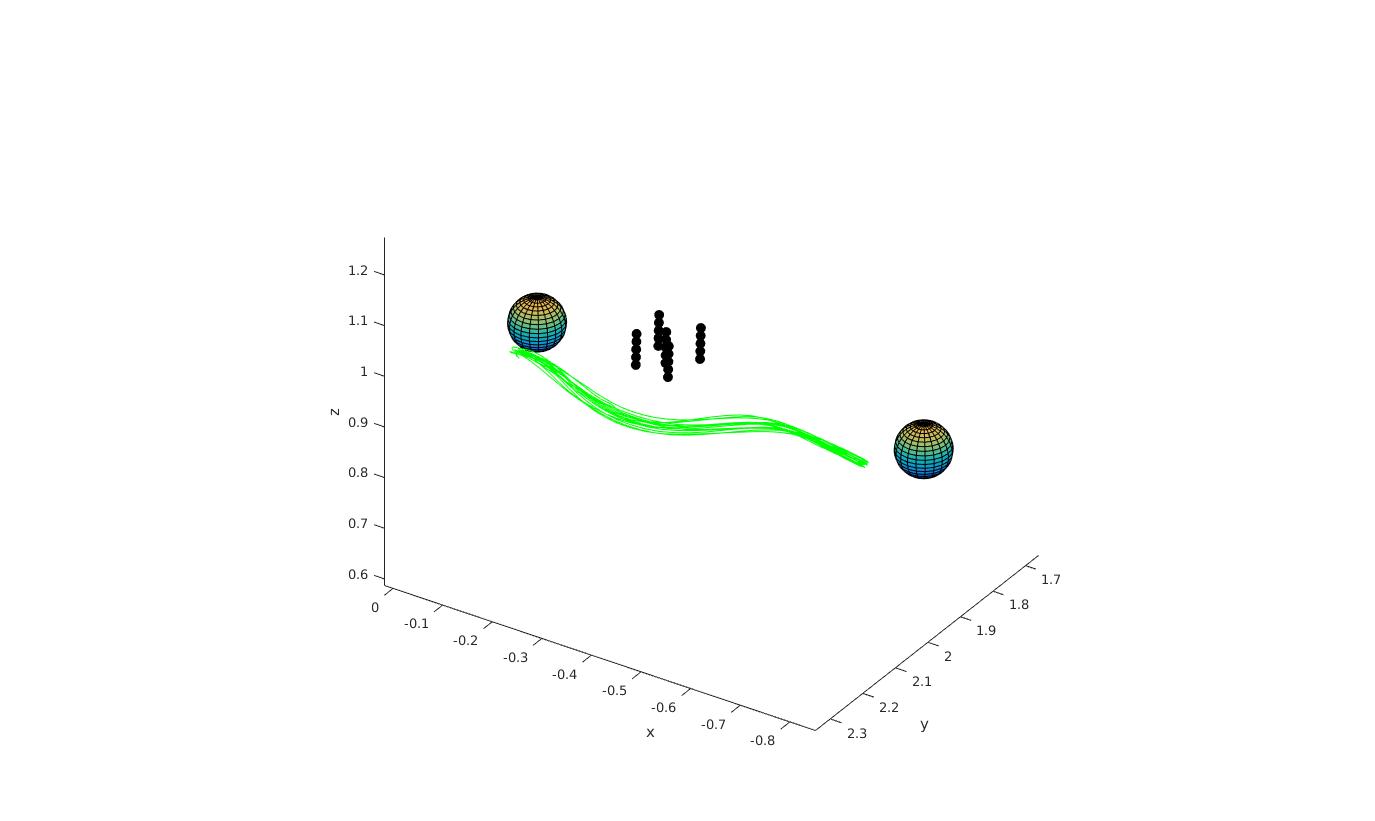}%
        \caption{Cube obstacle avoidance demonstrations.}
        \label{subfig:cube_all_good_samples_obs_avoid_119}
    \end{subfigure}
    \begin{subfigure}[b]{0.235\textwidth}
        \includegraphics[trim={9cm 2cm 9cm 2cm},clip,width=\textwidth]{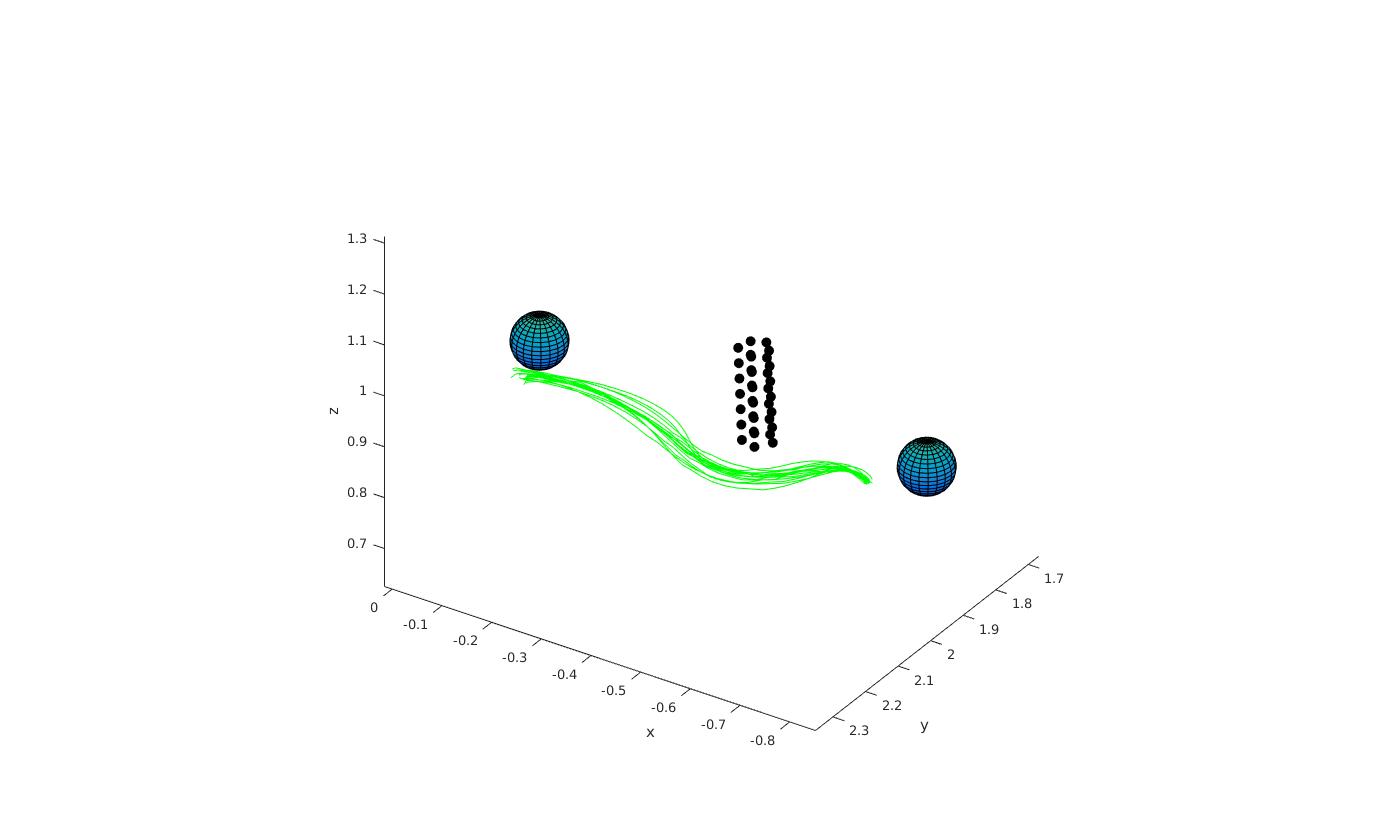}%
        \caption{Cylindrical obstacle avoidance demonstrations.}
        \label{subfig:cyl_all_good_samples_obs_avoid_194}
    \end{subfigure}
    \caption{Sample demonstrations. (b), (c), and (d) are a sample set of demonstrations for 1-out-of-40 settings.}
    \label{fig:demo_trajectories_display}
    \vspace{-0.2cm}
\end{figure}

To record human demonstrations we used a Vicon motion capture system
at 25 Hz sampling rate, with markers at the start position, goal
position, obstacle positions and the end-effector. These can be seen
in Figure \ref{fig:data_collection_setting}.
\GS{Giovanni: Done: Add a sentence about 1 baseline, 40 settings per
  obs, obs locations changed. Edit figure to say 1/1 40/120 etc.}
In total there are 40 different obstacle settings, each corresponding
to one obstacle position in the setup. We collected 21 demonstrations for the baseline (no
obstacle) behavior and 15 demonstrations of obstacle avoidance for
each obstacle settings with three different obstacles -- sphere, cube
and cylinder. From all baseline demonstrations, we learned one
baseline primitive, and all obstacle avoidance behaviors are assumed
to be a deviation of the baseline primitive, whose degree of deviation
is dependent on the obstacle setting. Some examples of the obstacle
avoidance demonstrations can be seen in Figure
\ref{fig:demo_trajectories_display}. Even though the Vicon setup only
tracked about 4-6 Vicon markers for each obstacle geometry, we
augmented the obstacle representation with more points to represent
the volume of each obstacle object.
\setcounter{subfigure}{0}
\subsection{Per setting experiments}
\begin{figure}
\includegraphics[scale=1]{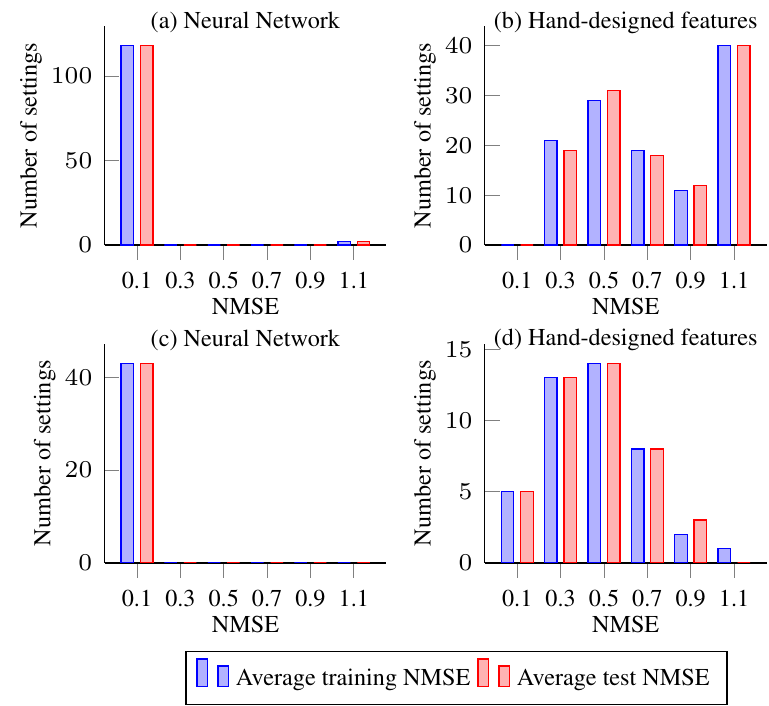}
\caption{Histograms describing the results of training and testing
  using a neural network (left plots) and model from
  \cite{rai2014learning} (plots to the right). (a) and (b) are average
  NMSE across all dimensions generated over the complete dataset. (c)
  and (d) are the NMSE over the dominant axis of demonstrations with
  obstacle avoidance.}
\label{fig:hist_all}
\vspace{-0.2cm}
\end{figure}
The per setting experiments were conducted on each setting separately.
We tried to incorporate demonstrations of near and far-away obstacles.
In total we test on 120 scenarios, comprised of 40 settings per
obstacle type (spheres, cylinder and cube).

A neural network was trained and unrolled over the particular setting
in question. For comparison, the model defined in
\cite{rai2014learning} was also trained on the same coupling term
target as the neural network. First, we evaluate and compare the
ability of the models to fit the training data and generalize to the
unseen test data (80/20 split). The consolidated results for these
experiments can be found in Figure~\ref{fig:hist_all}, where we show
the training and testing normalized mean square error (NMSE). The top
row (plots (a) and (b)) show results over all the scenarios (120) -
with the NMSE averaged across the 3 dimensions. The histogram shows,
for how many settings we achieved a particular training/testing NMSE.
As can be seen, when using the neural network, we achieved an NMSE of
0.1 or lower (for both training and testing data) in all scenarios -
indicating that the neural network indeed is flexible enough to fit
the data. The same is not true for the model of \cite{rai2014learning}
(plot b). However, a large portion of these settings have the obstacle
too far away such that there is no dominant axis of avoidance. The
model from \cite{rai2014learning} has a large training and testing
NMSE in such cases. We separated the demonstrations that have a
dominant axis of obstacle avoidance (43 scenarios) and show the
results for the dominant dimension of obstacle avoidance in plots (c)
and (d) of Figure~\ref{fig:hist_all}. As expected, the performance of
\cite{rai2014learning} features improves, but is still far behind the
performance of the neural network.
\begin{table}
\centering
\begin{tabular}{@{}rrrrrr@{}}\toprule
& \multicolumn{2}{c}{Distance} & \multicolumn{2}{c}{Distance} & {Number}\\
& \multicolumn{2}{c}{to goal} & \multicolumn{2}{c}{to obstacle} & {of hits}\\
\cmidrule{2-3} \cmidrule{4-5} 
& max & mean & min& mean &\\ \midrule
\ra{1.3}
\textbf{Initial demonstration} & 0.017 & 0.017 & -0.451	 & 4.992  & 4 \\
\textbf{Model from \cite{rai2014learning}} & 1.218 & 0.072 & -0.520 & 5.038 & 2 \\
\textbf{Neural Network} & 0.113 & 0.016 & 0.083 & 5.241 & 0 \\
\textbf{Human Demonstration} & 0.075 & 0.028 & 1.409 & 5.461 & 0 \\
\bottomrule
\end{tabular}
\caption{Results of the per setting experiments. Negative distance to obstacle implies a collision.}
\label{tab:results_per_setting}
\end{table}
\begin{table*}[ht!]
\centering
\begin{tabular}{@{}rrrrcrrcrrcr@{}}\toprule
&& \multicolumn{2}{c}{NMSE} & \phantom{abc} & \multicolumn{2}{c}{Distance to goal} & \phantom{abc} & \multicolumn{2}{c}{Distance to obstacle} & \phantom{abc} & {Number of hits}\\
\cmidrule{3-4} \cmidrule{6-7} \cmidrule{9-10}
&& train & test && max & mean && mean & min &&\\%
\midrule
\multirow{2}{*}{Sphere}& Baseline & - & - && 0.017 & 0.017 && 4.739 & -0.025 && 1\\%
& Unrolled & 0.155 & 0.152 && 0.152 & 0.018 && 5.063 & 0.433 && 0\\%
\multirow{2}{*}{Cube}& Baseline & - & - && 0.017 & 0.017 && 5.722 & -0.451 && 1\\%
& Unrolled & 0.164 & 0.159 && 0.145 & 0.015 && 5.964 & 1.045 && 0\\%
\multirow{2}{*}{Cylinder}& Baseline & - & - &&  0.017 & 0.017 && 4.514 & -0.280 && 2\\%
& Unrolled & 0.195 & 0.195 && 0.078 & 0.014 && 5.117 & 0.750 && 0\\%
\bottomrule
\end{tabular}
\caption{Results of the multi setting experiments. Negative distance to obstacle implies a collision.}
\label{tab:multi_setting_table}
\end{table*}

The features in \cite{rai2014learning} are unable to fit the human
data satisfactorily, as is illustrated in the high training NMSE. On
further study, we found that the issue with large regression weights
using Bayesian regression with ARD, as mentioned by the authors, can
be explained by a mismatch between the coupling term model used and
the target set. This also explains why they were not able to fit
coupling terms across settings.

The low training NMSE in Figure \ref{fig:hist_all} (a) and (c) show
the versatility of our neural network at fitting data very well per
setting. Low test errors showed that we were able to fit the data well
without over-fitting.

Note that the performance during unrolling for the same obstacle
setting can be different from the training demonstrations. When
unrolling, the DMP can reach states that were never explored during
training, and depending on the generalization of our model, we might
end up hitting the obstacle or diverge from our initial trajectory.
This brings up two points. One, we want to avoid the obstacle and two,
we want to converge to our goal in the prescribed time. We test both
methods on these two metrics, and the results are summarized in Table
\ref{tab:results_per_setting}. We compare the two learned coupling
term models to the baseline trajectory, as well as human demonstration
of obstacle avoidance. While the neural network never hits the
obstacle, the model from \cite{rai2014learning} hit the obstacle
twice. Likewise, the model from \cite{rai2014learning} does not always
converge to the goal, while the neural network always converges to the
goal. The mean distance to goal and mean distance from obstacle for
both methods are comparable to human demonstrations.
\subsection{Multiple setting experiments}
\begin{figure}
%\begin{figure}[h!]
    \centering
    \begin{subfigure}[b]{0.235\textwidth}
      \includegraphics[width=\textwidth]{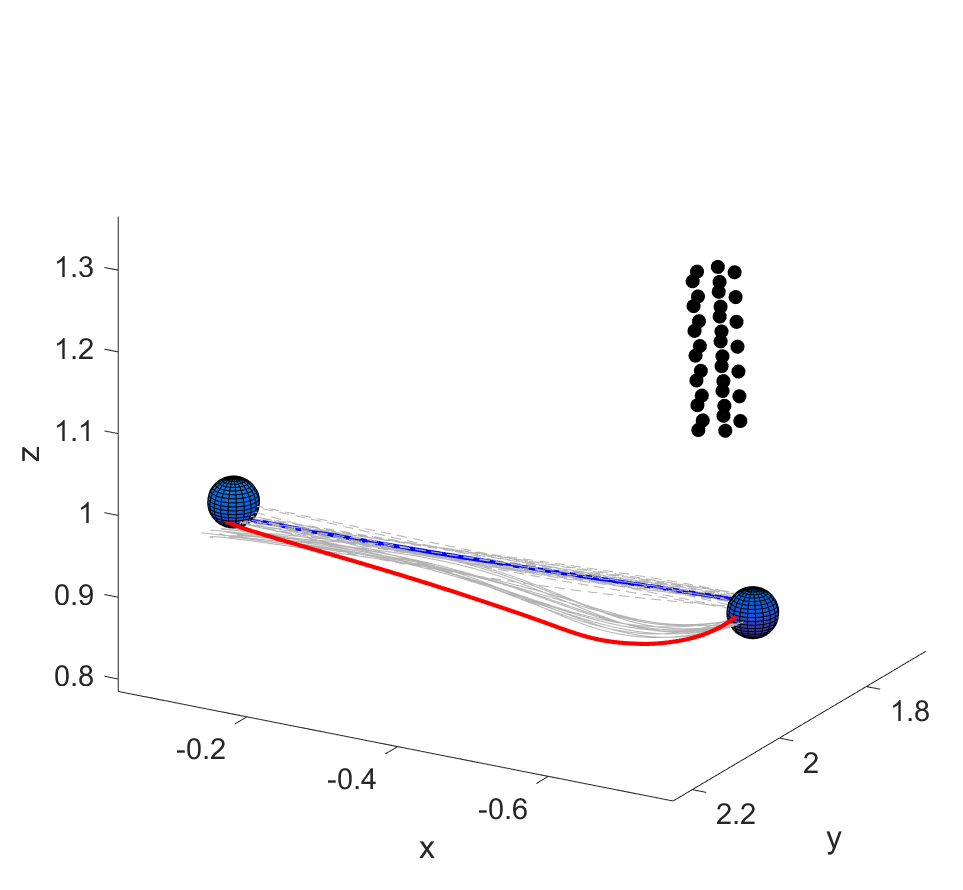}%
        \caption{Unroll on trained setting}
        \label{subfig:unroll}
    \end{subfigure}
    \begin{subfigure}[b]{0.235\textwidth}
        \includegraphics[width=\textwidth]{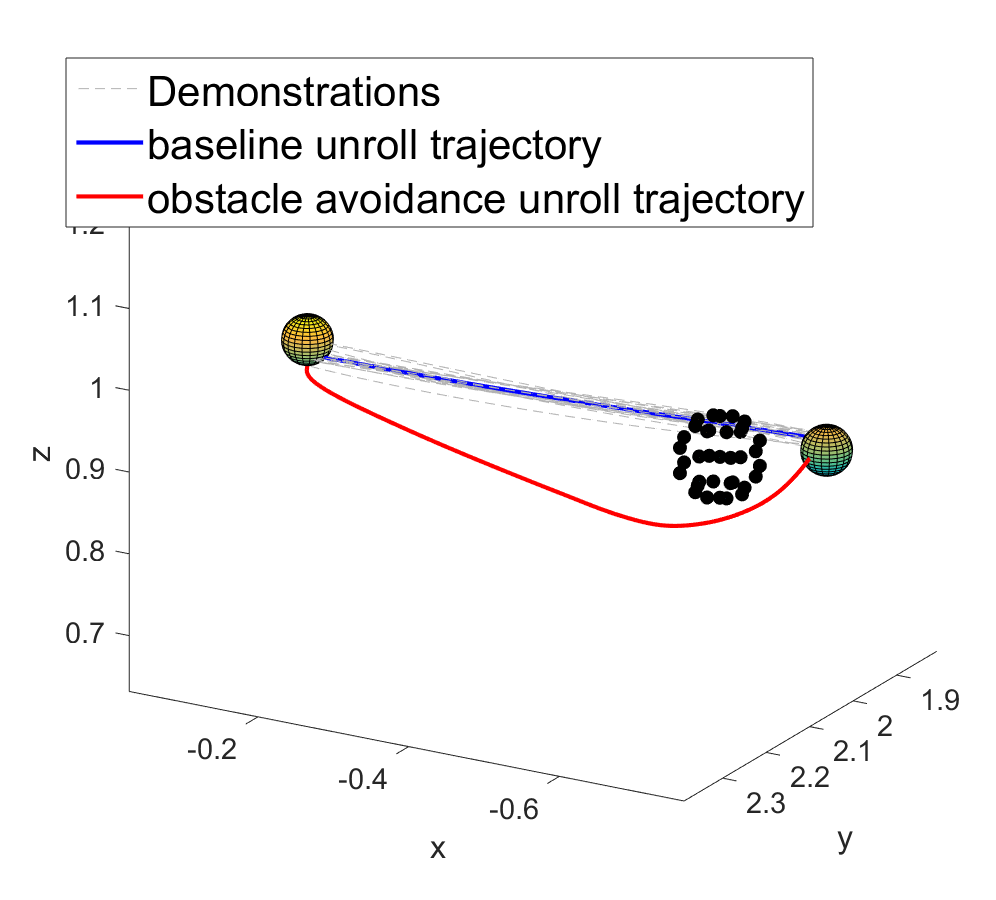}%
        \caption{Unroll on unseen setting}
        \label{subfig:unseen}
    \end{subfigure}
    \caption{Sample unrolled trajectories on trained and unseen settings.}
    \label{fig:unroll_trajectories}
\end{figure}

To test if our model generalizes across multiple settings of obstacle
avoidance, we train three neural networks over 40 obstacle avoidance
demonstrations per object. The results are summarized in Table
\ref{tab:multi_setting_table}. The neural network has relatively low
training and testing NMSE for the three obstacles. To test the
unrolling, each of the networks was used to avoid the 40 settings they
were trained on. As can be seen from columns 3 and 4, the unrolled
trajectories never hit an obstacle. They also converged to the goal in
all the unrolled examples. One example of unrolling on a trained
setting can be seen in Figure \ref{subfig:unroll}.
This shows that our neural network was able to learn coupling terms
across multiple settings and produce human-like, reliable obstacle
avoidance behavior, unlike previous coupling term models in
literature. When we trained our network across all three obstacles,
however, the performance deteriorated. We think this is because our
chosen inputs are very local in nature and to avoid multiple obstacles
the network needs a global input. In the future, we would like to use
features that can account for such global information across different
obstacles.
\subsection{Unseen setting experiments}
\AR{Akshara : Done: Why is unrolling different from training? Why can
  you unroll networks trained on cylinder on sphere obstacles, but not
  train over multiple obstacle?}
%\FM{how are the unseen settings generated?} 
To test generalization across unseen settings, we tested our trained
model on 63 unseen settings, initialized on a close
$7 \times 3 \times 3$ grid around the baseline trajectory. We
purposely created our unseen settings much harder than the trained
settings. Out of 63 settings, the baseline hit the obstacle in 35
demonstrations, as can be seen in Table~\ref{tab:unseen_setting}.
While our models were trained on spheres, cubes and cylinders, they
were all tested on spherical obstacles for simplicity. Please note
that while a model trained for cylinders can avoid spherical
obstacles, behaviorally the unrolled trajectory looks more like that
of cylindrical obstacle avoidance, than spherical.
%those from cylinder demonstrations, rather than spheres.
%
\begin{table}[ht]
\centering
\begin{tabular}{@{}rrrrrr@{}}\toprule
&  \multicolumn{2}{c}{Distance} & \multicolumn{2}{c}{Distance} & {Number}\\
&  \multicolumn{2}{c}{to goal} & \multicolumn{2}{c}{to obstacle} & {of hits}\\
\cmidrule{2-3} \cmidrule{4-5} 
&  max & mean & mean & min &\\%
\midrule
Initial & 0.017 & 0.017 & 0.095 & -0.918 & 35\\%
Sphere & 0.011 & 0.034 & 0.933 & -0.918 & 2 \\%
Cube & 0.021 & 0.119 & 2.235 & 1.172 & 0\\%
Cylinder & 0.033 & 0.120 & 1.704 & -0.103 & 1\\%
\bottomrule
\end{tabular}
\caption{Results of the unseen setting experiments. Negative distance to obstacle implies collision.}
\label{tab:unseen_setting}
\end{table}

As can be seen from Table~\ref{tab:unseen_setting}, our models were
able to generalize to unseen settings quite well. When trained on
sphere obstacle settings our approach hit the obstacle in 2 out of 63
settings, when trained on cylinder settings we hit it once, and when
trained on cube settings we never hit an obstacle. All the models
converged to the goal on all the settings. An example unrolling can be
seen in Figure~\ref{subfig:unseen}.
\FM{took out the tau invariance illustration - this is something for a
  journal paper. It's not a real experiment, it's more of a
  illustration, like the local coordinate transform figures in
  section 3.}
\subsection{Real robot experiment}
\begin{figure}[h]
    \centering
    \begin{subfigure}[t]{0.15\textwidth}
      \includegraphics[width=\textwidth]{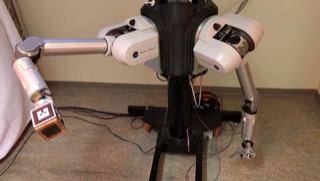}%
    \end{subfigure}
    \begin{subfigure}[t]{0.15\textwidth}
        \includegraphics[width=\textwidth]{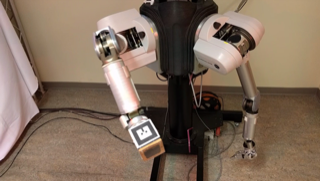}%
    \end{subfigure}
	\begin{subfigure}[t]{0.15\textwidth}
        \includegraphics[width=\textwidth]{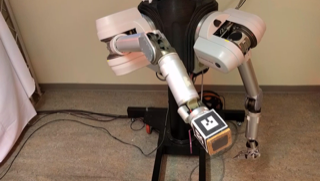}%
    \end{subfigure}
    \par\smallskip
    \begin{subfigure}[b]{0.15\textwidth}
      \includegraphics[width=\textwidth]{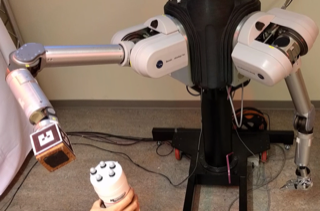}%
    \end{subfigure}
    \begin{subfigure}[b]{0.15\textwidth}
        \includegraphics[width=\textwidth]{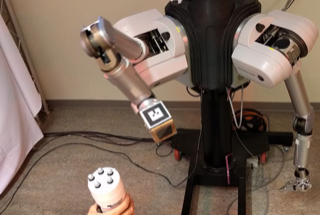}%
    \end{subfigure}
	\begin{subfigure}[b]{0.15\textwidth}
        \includegraphics[width=\textwidth]{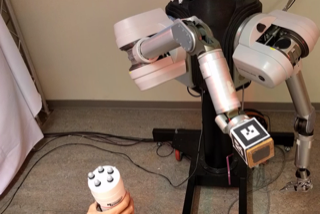}%
    \end{subfigure}
    \caption{Snapshots from our experiment on our real system. Here
      the robot avoids a cylindrical obstacle using a neural network
      that was trained over cylindrical obstacle avoidance
      demonstrations. See \url{https://youtu.be/hgQzQGcyu0Q} for the
      complete video.}
    \label{fig:hardware}
\end{figure}

%
%\FM{which one do we deploy?}
Finally we deploy the trained neural network on a 7 degree-of-freedom
Barrett WAM arm with 300 Hz real-time control rate, and test its
performance in avoiding obstacles. We again use Vicon objects tracked
in real-time at 25 Hz sampling rate to represent the obstacle. Some
snapshots of the robot avoiding a cylindrical obstacle using a neural
network trained on multiple cylindrical obstacles can be seen in
Figure \ref{fig:hardware}. Video can be seen in
\url{https://youtu.be/hgQzQGcyu0Q}.

%The Vicon data rate is 100 Hz, while the robot's control cycle rate is
%300 Hz, and we perform some window-averaging on the Vicon data, in
%order to minimize accidental data discontinuity from Vicon pose
%tracking error that might pose a danger to the robot's operation.

These are very promising results that show that a neural network with
intuitive features and physical constraints can generalize across
several settings of obstacle avoidance. It can avoid obstacles in
settings never seen before, and converge to the goal in a stable way.
This is a starting point for learning general feedback terms from data
that can generalize robustly to unseen situations.

\section{Discussion and Future Work}\label{sec:discussion}
In this paper, we introduce a general framework for learning feedback
terms from data, and test it on obstacle avoidance. We used a neural
network to learn a function that predicts the coupling term given
sensory inputs.
% , like end-effector position relative to obstacle,
% velocity, etc.
% The data for training this network was obtained from
% human demonstrations of obstacle avoidance on 120 settings, collected
% using a Vicon system.
Our results show that the neural network is able to fit the obstacle
avoidance demonstrations per setting as well as over multiple
settings. We also proposed to post-process the neural networks'
prediction based on physical constraints, that ensured that the
obstacle avoidance behavior was always stable and converged to the
goal in all the scenarios that we tested. When unrolled on trained
settings the DMP with online modulation via the neural network avoided
obstacles 100\% of the time, and when unrolled on unseen settings 98\%
of the time. We compared this work to an older coupling term model in
\cite{rai2014learning} and found our new results to be far more
impressive, in terms of fitting the data, as well as stability and
effectiveness in obstacle avoidance. We also deploy our approach on a
7 degree-of-freedom Barrett WAM arm using a Vicon system and it
successfully avoids obstacles.

However, when training across obstacles, the performance of the neural
network deteriorates. This could be because generalization across
different obstacles needs some global information about the obstacle.
In the future, we would like to add some global inputs to try and
learn a model across obstacles. Eventually, we would also like to
learn coupling terms for tasks other than obstacle avoidance and see
the validity of our approach in other problems. Our post-processing
too, is focused on obstacle avoidance right now. For more general
problems, we might need to add other constraints, for example torque
saturation to ensure stable and safe behavior.

The choice of using a neural network was partially influenced by our
long term vision of a general approach to learning feedback terms. For
instance, it would be interesting to learn a more complex network that
takes raw sensor information -- such as visual feedback -- as input,
requiring even lesser human design.

This paper is a step towards automatically learning feedback terms
from data and producing safe, generalizable coupling terms that can
modify the current plan reactively without re-planning. We are trying
to minimize human-designed inputs and tuned parameters in our control
approach. Our promising results establish its validity for obstacle
avoidance, but how well this performance can be transferred to other
tasks still remains to be seen.

\addtolength{\textheight}{-12cm}   % This command serves to balance the column lengths

%%%%%%%%%%%%%%%%%%%%%%%%%%%%%%%%%%%%%%%%%%%%%%%%%%%%%%%%%%%%%%%%%%%%%%%%%%%%%%%%

%%%%%%%%%%%%%%%%%%%%%%%%%%%%%%%%%%%%%%%%%%%%%%%%%%%%%%%%%%%%%%%%%%%%%%%%%%%%%%%%



% \section*{ACKNOWLEDGMENT}

% The preferred spelling of the word ÒacknowledgmentÓ in America is without an ÒeÓ after the ÒgÓ. Avoid the stilted expression, ÒOne of us (R. B. G.) thanks . . .Ó  Instead, try ÒR. B. G. thanksÓ. Put sponsor acknowledgments in the unnumbered footnote on the first page.

%%%%%%%%%%%%%%%%%%%%%%%%%%%%%%%%%%%%%%%%%%%%%%%%%%%%%%%%%%%%%%%%%%%%%%%%%%%%%%%%

\bibliographystyle{IEEEtran}
\bibliography{references}

\end{document}